\documentclass[letterpaper, 10 pt, conference]{ieeeconf}  %

\IEEEoverridecommandlockouts                              %

\usepackage{graphicx} %
\usepackage{amsmath} %
\usepackage{amssymb}  %
\usepackage{color}
\usepackage[table]{xcolor}
\usepackage{colortbl}
\usepackage{adjustbox}
\usepackage{siunitx}
\usepackage{subcaption}
\usepackage{hhline}
\usepackage{siunitx}

\usepackage{footmisc} %

\makeatletter
\let\NAT@parse\undefined
\makeatother
\usepackage[bookmarks=false]{hyperref}

\usepackage{dblfloatfix}

\usepackage[capbesideposition=inside,
facing=yes,capbesidesep=quad]{floatrow}
\usepackage[textsize=scriptsize]{todonotes}
\usepackage{booktabs}
\usepackage{multirow}
\let\labelindent\relax
\usepackage{enumitem}
\usepackage{fancyhdr}
\newcommand{\initial}[1]{}

\title{\LARGE \bf
Efficient RGB-D Semantic Segmentation for Indoor Scene Analysis
}

\author{Daniel Seichter, Mona Köhler, Benjamin Lewandowski, Tim Wengefeld and Horst-Michael Gross%
\thanks{Authors are with Neuroinformatics and Cognitive Robotics Lab, Technische Universit\"at Ilmenau, 98693 Ilmenau, Germany. {\tt\small daniel.seichter@tu-ilmenau.de}}%
\thanks{This work has received funding from the German Federal Ministry of Education and Research (BMBF) to the project MORPHIA (grant agreement no. 16SV8426) and from the Carl Zeiss Foundation to the project E4SM (grant agreement no. P2017-01-005).}%
}

\begin{document}
\bstctlcite{IEEEexample:BSTcontrol}
\maketitle

\thispagestyle{fancy}
\fancyhf{}
\fancyhead[C]{%
    \footnotesize%
    \vspace{-15mm}%
    \textcolor{gray}{%
        © 2021 IEEE.  Personal use of this material is permitted.  Permission from IEEE must be obtained for all other uses, in any current or future media, including reprinting/republishing this material for advertising or promotional purposes, creating new collective works, for resale or redistribution to servers or lists, or reuse of any copyrighted component of this work in other works.
    }%
}
\renewcommand{\headrulewidth}{0pt}

\pagestyle{empty}

\begin{abstract}
Analyzing scenes thoroughly is crucial for mobile robots acting in different environments. 
Semantic segmentation can enhance various subsequent tasks, such as (semantically assisted) person perception, (semantic) free space detection, (semantic) mapping, and (semantic) navigation.
In this paper, we propose an efficient and robust RGB-D segmentation approach that can be optimized to a high degree using NVIDIA TensorRT and, thus, is well suited as a common initial processing step in a complex system for scene analysis on mobile robots.
We show that RGB-D segmentation is superior to processing RGB images solely and that it can still be performed in real time if the network architecture is carefully designed.
We evaluate our proposed Efficient Scene Analysis Network (ESANet) on the common indoor datasets NYUv2 and SUNRGB-D and show that we reach state-of-the-art performance while enabling faster inference.
Furthermore, our evaluation on the outdoor dataset Cityscapes shows that our approach is suitable for other areas of application as well. 
Finally, instead of presenting benchmark results only, we also show qualitative results in one of our indoor application scenarios.
\end{abstract}

\section{Introduction}
\label{sec:introduction}
Semantic scene perception and understanding is essential for mobile robots acting in various environments.
In our research projects, covering public environments from supermarkets~\cite{Gross-IROS-2009, Lewandowski-ROMAN-2020} to hospitals~\cite{Gross-ICRA-2017,Trinh-ROMAN-2020} and domestic applications~\cite{Gross-IROS-2015,Gross-ICRA-2019}, our robots need to perform several tasks in parallel such as obstacle avoidance, semantic mapping, navigation to semantic entities, and person perception. 
Most of the tasks require to be handled in real time given limited computing and battery capabilities.
Hence, an efficient and shared initial processing step can facilitate subsequent tasks. 
Semantic segmentation is well suited for such an initial step, as it provides precise pixel-wise information that can be used for numerous subsequent tasks.

\begin{figure}[t!]
    \vspace{2mm}
	\centering
	\includegraphics[width=\columnwidth]{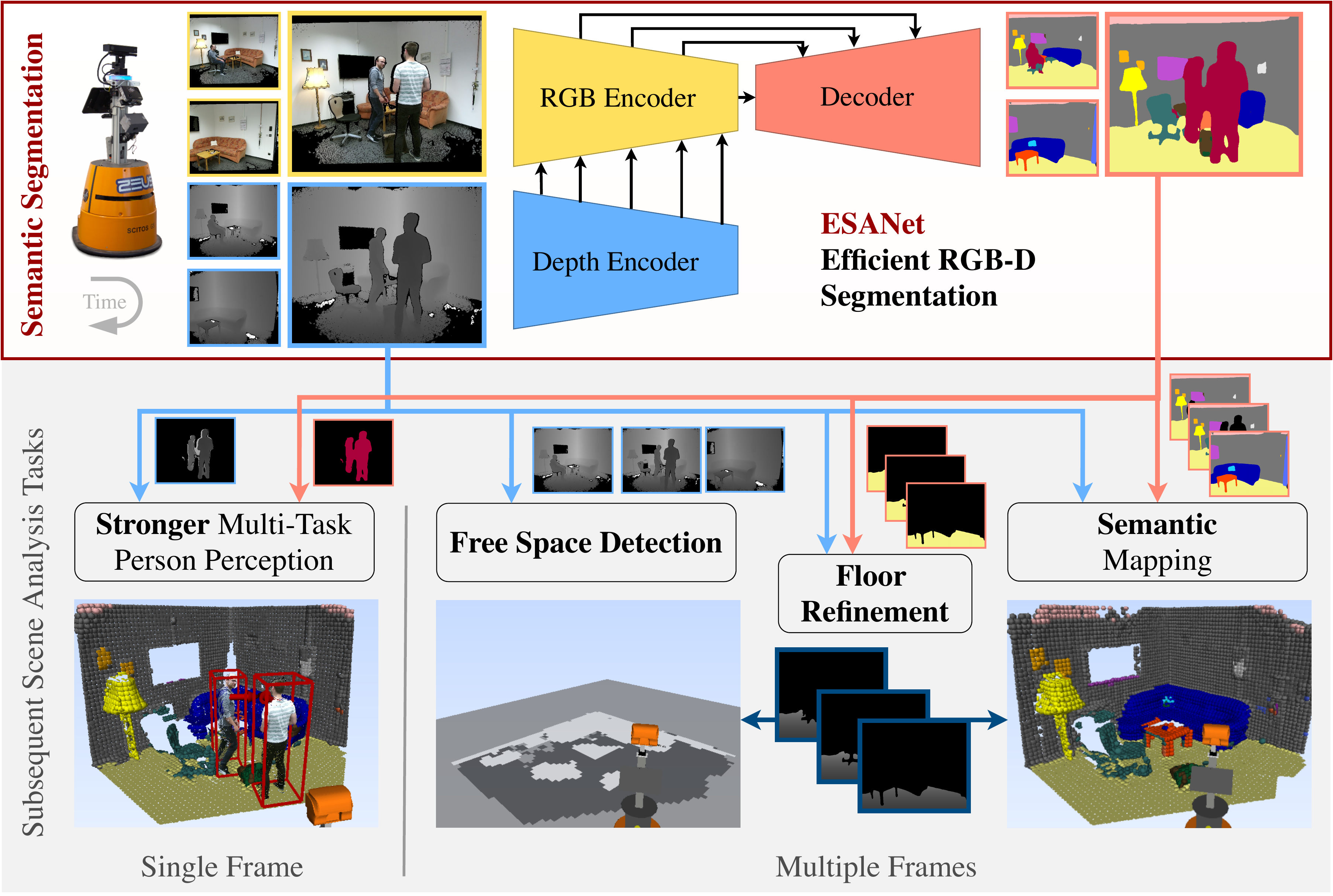}
	\caption{
	Our proposed efficient RGB-D segmentation approach can serve as a common preprocessing step for subsequent tasks such as person perception, free space detection to avoid even small obstacles below the laser, or semantic mapping.}
	\label{fig:eyecatcher}
\end{figure}

In this paper, we propose an efficient and robust encoder-decoder-based semantic segmentation approach that can be embedded in complex systems for semantic scene analysis such as shown in Fig.~\ref{fig:eyecatcher}.
The segmentation output enriches the robot's visual perception and facilitates subsequent processing steps by providing individual semantic masks.
For our person perception~\cite{Seichter-IROS-2020}, computations can be restricted to image regions segmented as person, instead of processing the entire image.
Furthermore, the floor class indicates free space that can be used for inpainting invalid depth pixels as well as serves as additional information for avoiding even small obstacles below the laser.
For mapping~\cite{Einhorn-ECMR-2013}, we can include semantics and ignore image regions segmented as dynamic classes such as person. 

Our segmentation approach relies on both RGB and depth images as input.
Especially in indoor environments, cluttered scenes may impede semantic segmentation. 
Incorporating depth images can alleviate this effect by providing complementary geometric information, as shown in \cite{FuseNet, RedNet, ACNet}.
In contrast to processing RGB images solely, this design decision comes with some additional computational cost.
However, in this paper, we show that two carefully designed shallow encoder branches (one for RGB and one for depth data) can achieve better segmentation performance while still enabling faster inference~(application of network) than a single deep encoder branch for RGB images solely.
Moreover, our Efficient Scene Analysis Network (ESANet) enables much faster inference than most other RGB-D segmentation methods, while performing on par or even better as shown by our experiments.

We evaluate our ESANet on the commonly used indoor datasets NYUv2~\cite{NYUv2} and SUNRGB-D~\cite{SUNRGBD} and further present qualitative results in our indoor application.
Instead of only focusing on mean intersection over union (mIoU) as evaluation metric for benchmarking, we also strive for fast inference on embedded hardware.
However, rather than reporting the inference time on high-end GPUs, we measure inference time on our robot's NVIDIA Jetson AGX Xavier. 
We designed our network architecture such that it can be executed as a single optimized graph using NVIDIA TensorRT.
Moreover, by evaluating on the outdoor dataset Cityscapes~\cite{Cityscapes}, we show that our approach can also be applied to other areas of application.

\noindent The main contributions of this paper are
\begin{itemize}[leftmargin=5mm]
	\item an efficient RGB-D segmentation approach, which can serve as initial processing step to facilitate subsequent scene analysis tasks and is characterized by:
	\begin{itemize}
	    \item a carefully designed architecture that can be optimized to a high degree using NVIDIA TensorRT and, thus, enables fast inference
    	\item an efficient ResNet-based encoder that utilizes a modified basic block that is computationally less expensive while achieving higher accuracy
    	\item a decoder that utilizes a novel learned upsampling
	\end{itemize}
	\item a detailed ablation study to the fundamental parts of our approach and their impact on segmentation performance and inference time
	\item qualitative results in a complex system for robotic scene analysis proving the applicability and robustness. %
\end{itemize}
Our code as well as the trained networks are publicly available at: \href{https://github.com/TUI-NICR/ESANet}{\small \url{https://github.com/TUI-NICR/ESANet}}

\section{Related Work}
\label{sec:related_work}
Common network architectures for semantic segmentation follow an encoder-decoder design.
The encoder extracts semantically rich features from the input and performs downsampling to reduce computational effort. 
The decoder restores the input resolution and, finally, assigns a semantic class to each input pixel.

\subsection{RGB-D Semantic Segmentation}
\label{sec:related_work:rgbd}
Depth images provide complementary geometric information to RGB images and, thus, improve segmentation~\cite{FuseNet, RedNet}. %
However, incorporating depth information into RGB segmentation architectures is challenging as depth introduces deviating statistics and characteristics from another modality.

In \cite{3D-Geometry-Aware}, depth information is used to project the RGB images into a 3D space.
However, processing the resulting 3D data, leads to significantly higher computational complexity.
\cite{2.5DConvolutionRGBD, Malleable2.5D, DepthAwareCNN, SGNet, 3DN} design specifically tailored convolutions, taking into account depth information.
Nevertheless, these modified convolutions often lack optimized implementations and, thus, are slow and not applicable for real-time segmentation on embedded hardware.

The majority of approaches for RGB-D segmentation~\cite{FuseNet, CouplingTwoStream, RedNet, ACNet, SA-Gate, RDFNet, SSMA, MMAF-Net} simply use two branches, one for RGB and one for depth data and fuse the feature representations later in the network.
This way, each branch can focus on extracting modality-specific features, such as color and texture from RGB images and geometric, illumination-independent features from depth images.
Fusing these modality-specific features leads to stronger feature representations.
Instead of fusing only low-level or high-level features, \cite{FuseNet} shows that the segmentation performance increases if the features are fused at multiple stages.
Typically, the features are fused once at each resolution stage with the last fusion at the end of both encoders.
Using only one decoder for the combined features reduces the computational effort.
FuseNet~\cite{FuseNet} and RedNet~\cite{RedNet} fuse the depth features into the RGB encoder, which follows the intuition that the semantically richer RGB features can be further enhanced using complementary depth information.
SA-Gate \cite{SA-Gate} combines RGB and depth features and fuses the recalibrated features back into both encoders.
In order to make the two encoders independent of each other, ACNet~\cite{ACNet} uses an additional, virtual, third encoder that obtains modality-specific features from the two encoders and processes the combined features.
Instead of fusing in the encoder, the modality-specific features can also be used to refine the features in the common decoder via skip connections as in RDFNet~\cite{RDFNet}, SSMA~\cite{SSMA} and MMAF-Net~\cite{MMAF-Net}.

However, none of the aforementioned methods focus on efficient RGB-D segmentation for embedded hardware. 
Using deep encoders such as ResNets with 50, 101 or even 152 layers results in high inference times and, therefore, makes them inappropriate for deploying to mobile robots.

\subsection{Efficient Semantic Segmentation}
\label{sec:related_work:efficient}
In contrast to RGB-D segmentation, recent RGB approaches \cite{ERFNet, LEDNet, DABNet, EDANet, SwiftNet, BiSeNet, ESPNetv2} also address reducing computational complexity to enable real-time segmentation.
Most efficient segmentation approaches propose specifically tailored network architectures, which reduce both the number of operations and parameters to enable faster inference while still retaining good segmentation performance.
Approaches, such as ERFNet~\cite{ERFNet}, LEDNet~\cite{LEDNet}, or DABNet~\cite{DABNet} introduce efficient encoder blocks by replacing expensive $3{\times}3$ convolutions with more light-weight variants such as factorized, grouped, or depth-wise separable convolutions.
Nevertheless, although requiring more operations and memory, SwiftNet~\cite{SwiftNet} and BiSeNet~\cite{BiSeNet} are still faster than many other methods, while achieving higher segmentation performance by simply using a pretrained ResNet18 as encoder.
This can be deduced to utilizing early and high downsampling in the encoder, a light-weight decoder, and using standard $3{\times}3$ convolutions, which are currently implemented more efficiently than grouped or depth-wise convolutions and have large representational power.

Following SwiftNet and BiSeNet, our approach also uses a ResNet-based encoder.
However, in order to further reduce inference time, we exchange the basic block in all ResNet layers with a more efficient block, based on factorized convolutions.

\begin{figure*}[t!]%
    \vspace{1mm}
        \resizebox{0.98\textwidth}{!}{%
            \begin{tikzpicture}[every node/.style={inner sep=0,outer sep=0}]%
                \node[anchor = south] at(0, -0.5){
                    \includegraphics[scale=0.5]{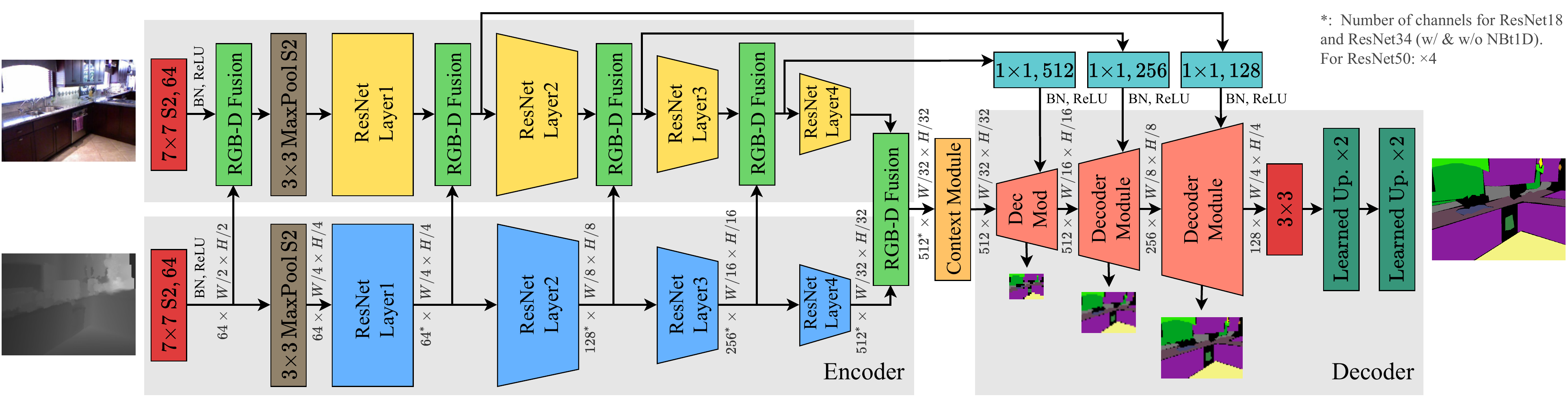}
                };%
                \definecolor{rgbd_fusion}{HTML}{6DD46A}
                \node[anchor = south west] at(-9.7, -5.8){
                    \includegraphics[scale=0.5]{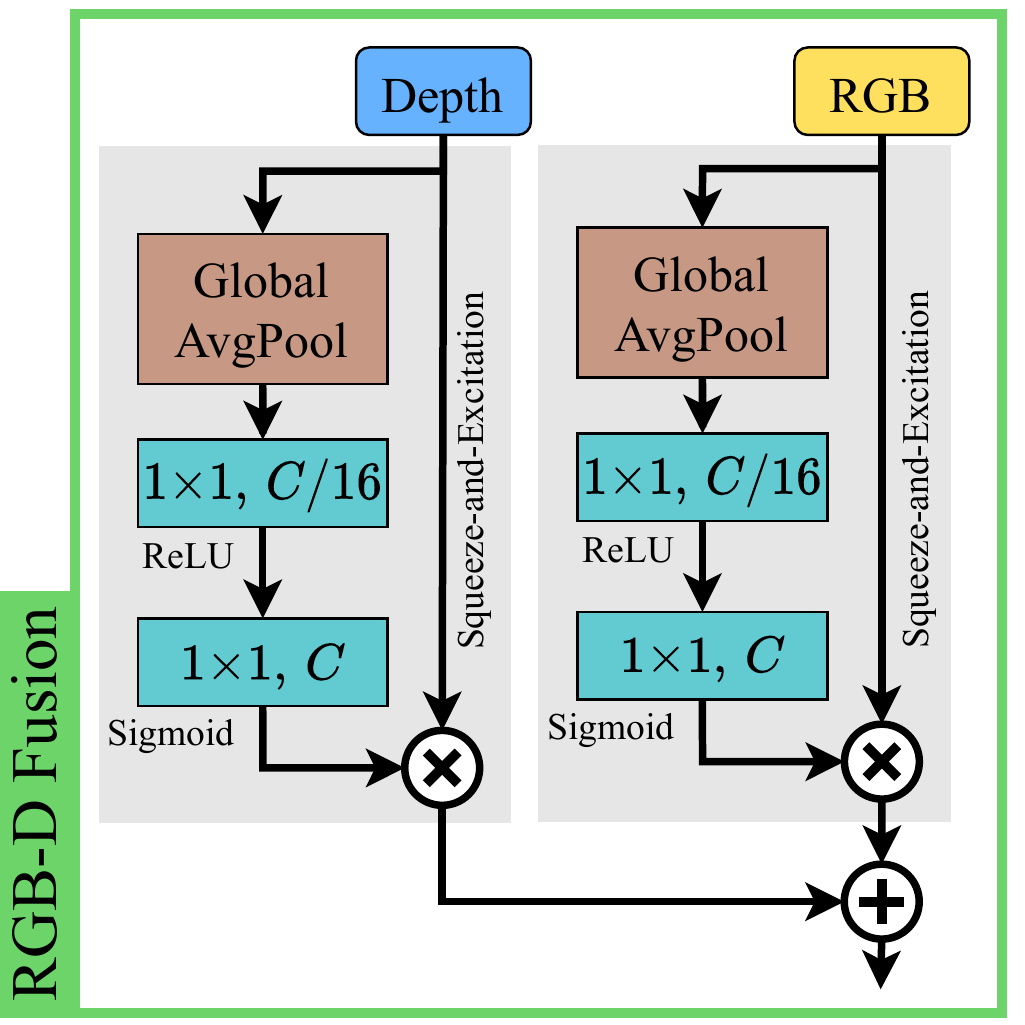}
                };%
                \definecolor{context_module}{HTML}{FFC266}
                \node[anchor = south west] at(-4.3, -5.8){
                    \includegraphics[scale=0.5]{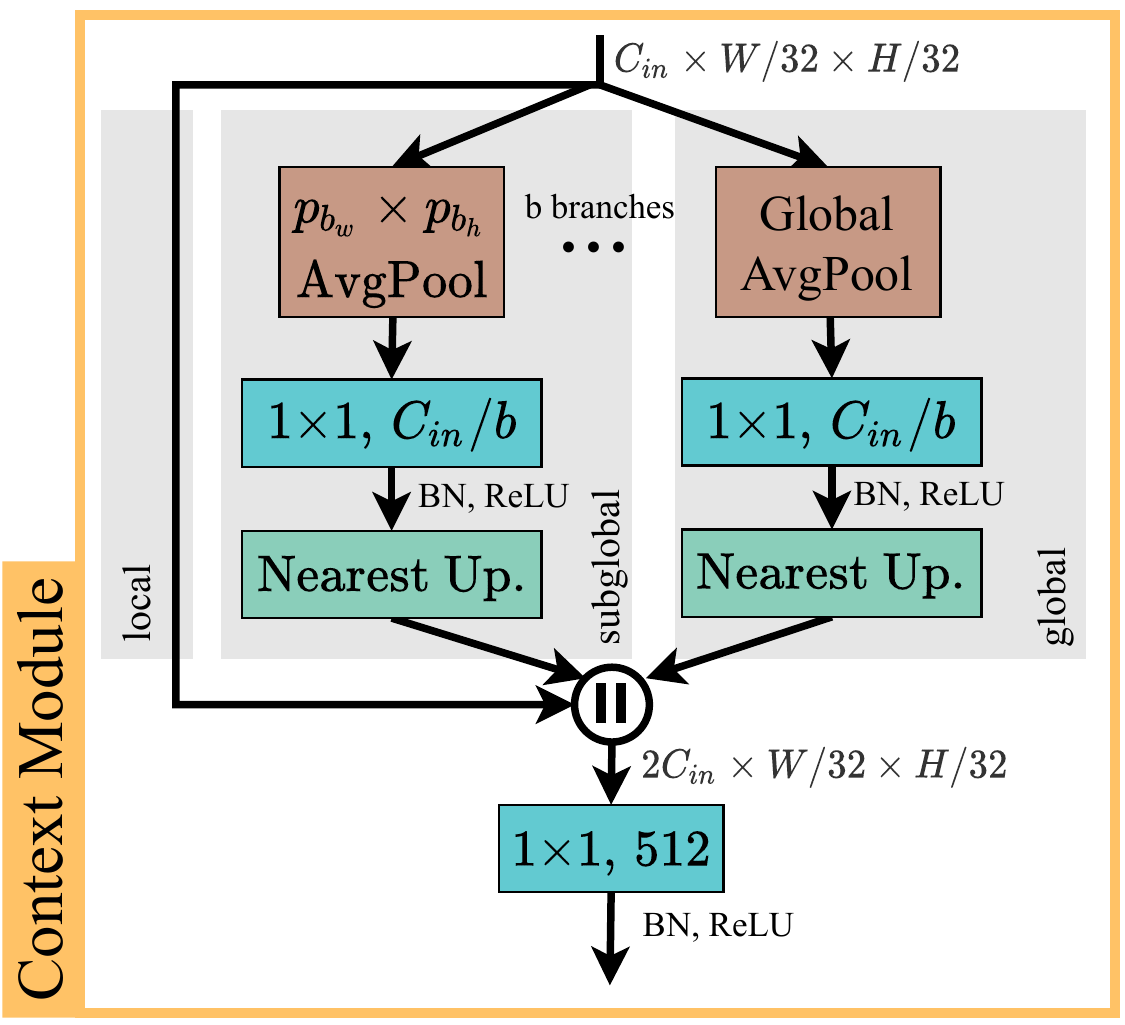}
                };%
                \definecolor{decoder_module}{HTML}{FF8573}
                \node[anchor = south west] at(1.7, -5.8){
                    \includegraphics[scale=0.5]{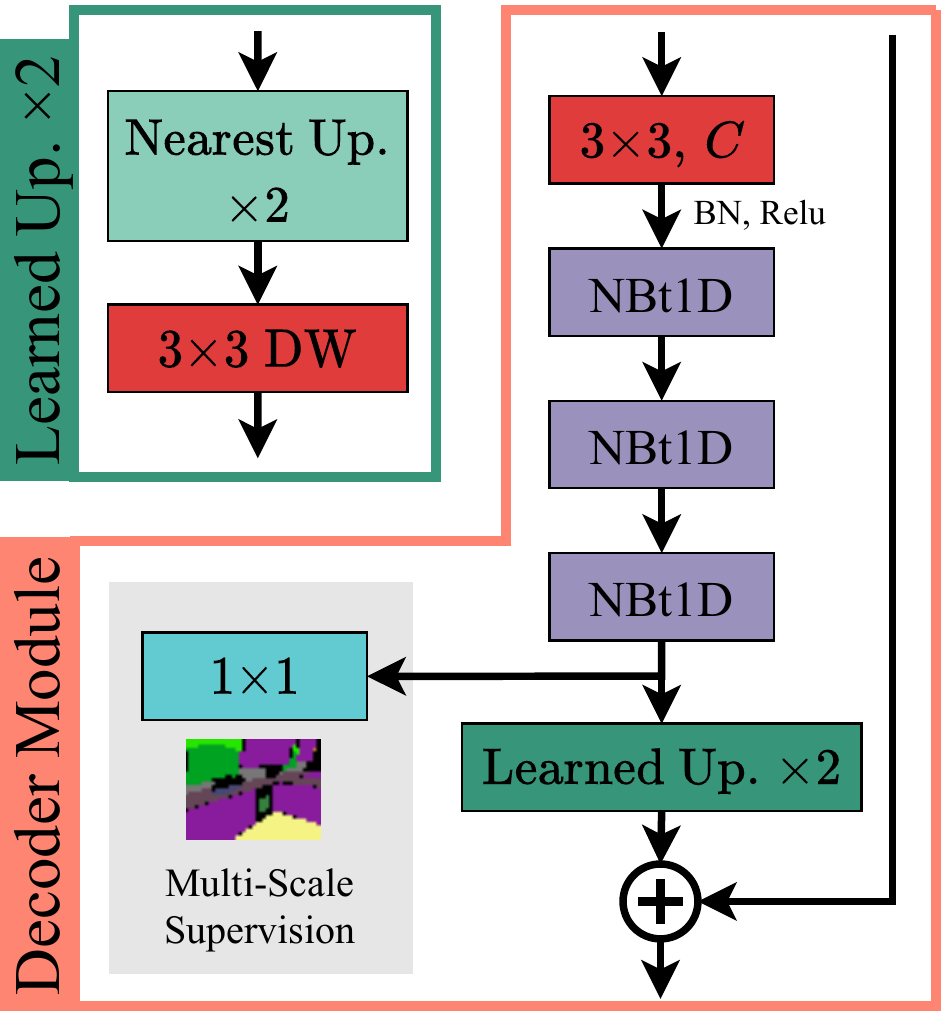}
                };%
                \definecolor{nbt1d}{HTML}{9B91BD}
                \node[anchor = south west] at(6.7, -5.8){
                    \includegraphics[scale=0.5]{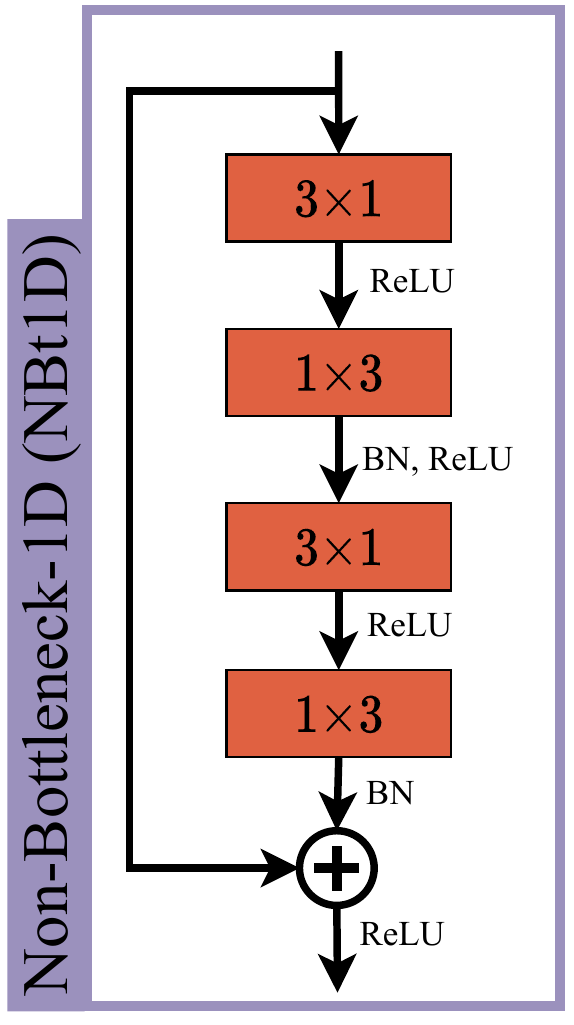}
                };%
                \node[anchor = south west] at (-9.7, -6.4){
                    \scriptsize
                    \definecolor{gray_legend}{RGB}{70,70,70}%
                    \textcolor{gray_legend}{
                        Legend:  \fbox{$k_w{\times}k_h, C$} : Convolution with kernel size $k_w{\times}k_h$ and $C$ output channels, S2: Stride 2, BN: Batch Normalization, Up.: Upsampling, DW: Depthwise, \hspace{3.4mm}: Concatenation
                    }
                };%
                \node[anchor = south west] at(7.05, -6.3){
                    \includegraphics[scale=0.35]{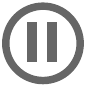}
                };%
            \end{tikzpicture}%
        }%
    
    \vspace{-3mm}
    \caption{Overview of our proposed ESANet for efficient RGB-D segmentation (top) and specific network parts (bottom).\\[-4.5mm]}
    \label{fig:model}
\end{figure*}

\section{Efficient RGB-D Segmentation}
\label{sec:approach}
The architecture of our Efficient Scene Analysis Network~(ESANet) for RGB-D semantic segmentation is depicted in Fig.~\ref{fig:model} (top). 
It is inspired by the RGB segmentation approach SwiftNet~\cite{SwiftNet}, i.e., a shallow encoder with a pretrained ResNet18 backbone and large downsampling, a context module similar to the one in PSPNet~\cite{PSPNet}, a shallow decoder with skip connections from the encoder, and final upsampling by a factor of 4.
However, SwiftNet does not incorporate depth information at all. 
Therefore, our ESANet uses an additional encoder for depth data.
This depth encoder extracts complementary geometric information that is fused into the RGB encoder at several stages using an attention mechanism.
Furthermore, both encoders use a revised architecture enabling faster inference.
The decoder is comprised of multiple modules, each is upsampling the resulting feature maps by a factor of 2 and is refining the features using convolutions as well as by incorporating encoder features.
Finally, the decoder maps the features to the classes and rescales the class mapping to the input resolution.

Our entire network features simple components implemented in PyTorch~\cite{pytorch}. 
We do not use complex structures or specifically tailored operations as these are often incompatible for converting to ONNX~\cite{onnx} or NVIDIA TensorRT and, thus, result in slower inference time.

In the following, we explain each part of our network design in detail as well as its motivation.
Fig.~\ref{fig:model} (bottom) depicts the exact structure of our network modules.

\subsection{Encoder}
\label{sec:approach:encoder}
The RGB and depth encoder both use a ResNet architecture~\cite{ResNet} as backbone.
For efficiency reasons, we do not replace strided convolutions by dilated convolutions as in PSPNet~\cite{PSPNet} or DeepLabv3~\cite{DeepLabv3}.
Thus, the resulting feature maps at the end of the encoder are 32 times smaller than the input image. 
For a trade-off between speed and accuracy, we use ResNet34 but also show results for ResNet18 and ResNet50.
We replace the basic block in each layer of ResNet18 and ResNet34 with a spatially factorized version. 
More precisely, each $3{\times}3$ convolution is replaced by a $3{\times}1$ and a $1{\times}3$ convolution with a ReLU in-between.
The so-called \textit{Non-Bottleneck-1D-Block} (NBt1D) is depicted in Fig.~\ref{fig:model} (violet) and was initially proposed in ERFNet~\cite{ERFNet} for another network architecture. 
In our experiments, we show that this block can also be used in ResNet and simultaneously reduces inference time and increases segmentation performance. 

\subsection{RGB-D Fusion}
At each of the five resolution stages in the encoders (see Fig.~\ref{fig:model}), depth features are fused into the RGB encoder.
The features from both modalities are first reweighted with a Squeeze and Excitation (SE) module \cite{SENet} and then summed element-wisely, as shown in Fig.~\ref{fig:model}~(light green).
Using this channel attention mechanism, the model can learn which features of which modality to focus on and which to suppress, depending on the given input.
In our experiments, we show that this fusion mechanism notably improves segmentation.

\subsection{Context Module}
Due to the limited receptive field of ResNet~\cite{PSPNet}, we additionally incorporate context information by aggregating features at different scales using several branches in a context module similar to the Pyramid Pooling Module in PSPNet~\cite{PSPNet} (see Fig.~\ref{fig:model}~orange).
Since NVIDIA TensorRT only supports pooling with fixed sizes, we carefully designed the context module such that the pooling sizes are always a factor of the input resolution of the context module and no adaptive pooling is required.
Note that, depending on the image resolution of the respective dataset, the number of existing factors and, thus, the branches $b$ and pooling sizes~$p_{b_{w}}{\times}p_{b_{h}}$ differ.
Our experiments show that this additional context module improves segmentation.
\begin{figure*}[!b]
    \vspace{-4mm}
    \centering
    \begin{tikzpicture}
        \node[anchor = south] at(-3.8,0){
            \includegraphics[width=0.18\textwidth]{img/upsampling/0761.png}
        };%
        \node[anchor = north] at(-3.8,0.1){\footnotesize{RGB image}};%
        \node[anchor = south] at(0,0){
            \includegraphics[width=0.18\textwidth]{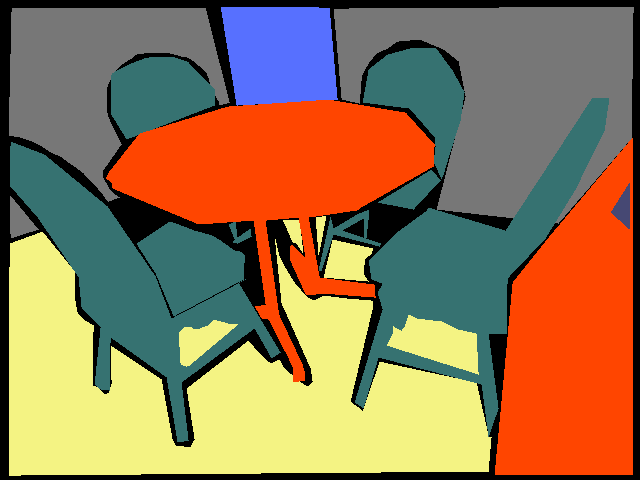}
        };%
        \node[anchor = south] at(-0.2,2.1){\tiny{door}};%
        \node[anchor = south, color=white] at(1.1,2.0){\tiny{wall}};%
        \node[anchor = south, color=white] at(0.6,1.05){\tiny{chair}};%
        \node[anchor = south] at(-0.3,1.55){\tiny{table}};%
        \node[anchor = south] at(1.27,0.9){\tiny{table?}};%
        \node[anchor = south] at(0,0.15){\tiny{floor}};%
        \node[anchor = north] at(0,0.1){\footnotesize{Ground Truth}};%
        \node[anchor = south] at(3.5,0){
            \includegraphics[width=0.18\textwidth]{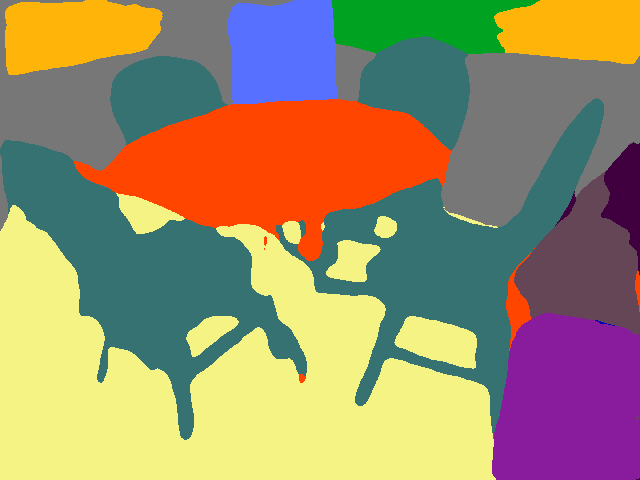}
        };%
        \node[anchor = south] at(2.25,2.15){\tiny{picture}};%
        \node[anchor = south] at(4.0,2.25){\tiny{window}};%
        \node[anchor = south,color = white] at(4.77,0.97){\tiny{counter}};%
        \node[anchor = south,color = white] at(4.75,0.2){\tiny{cabinet}};%
        \node[anchor = north] at(3.5,0.1){\footnotesize{\bf Learned Up. (ours)}};%
        \node[anchor = south] at(7,0){
            \includegraphics[width=0.18\textwidth]{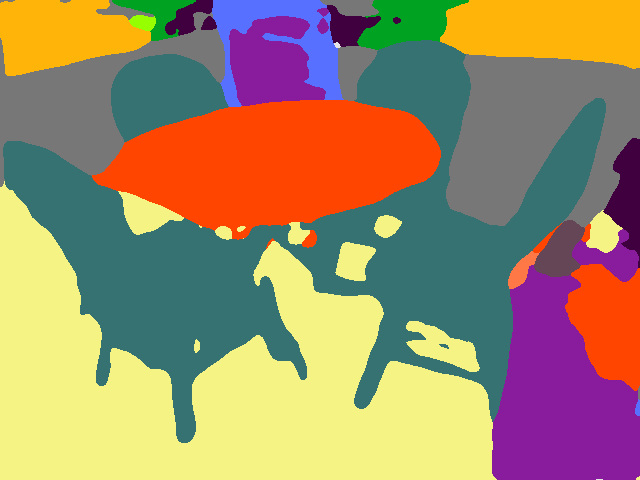}
        };%
        \node[anchor = north] at(7,0.1){\footnotesize{Bilinear Up.}};%
        \node[anchor = south] at(10.5,0){
            \includegraphics[width=0.18\textwidth]{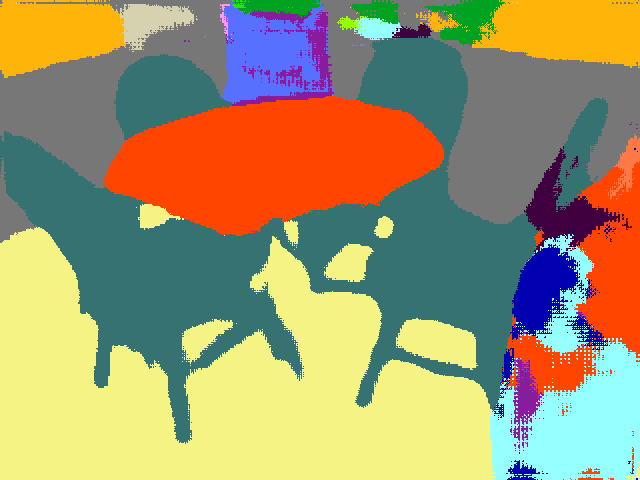}
        };%
        \node[anchor = south] at(11.85,0.15){\tiny{bed}};%
        \node[anchor = north] at(10.5,0.1){\footnotesize{Transp. Conv. (ACNet~\cite{ACNet})}};%
    \end{tikzpicture}
    
    \vspace{-3mm}
    \caption{Qualitative comparison of upsampling methods on NYUv2 test set (same colors as in Fig.~\ref{fig:eyecatcher} and Fig.~\ref{fig:application}).}
    \label{fig:upsampling}
\end{figure*}
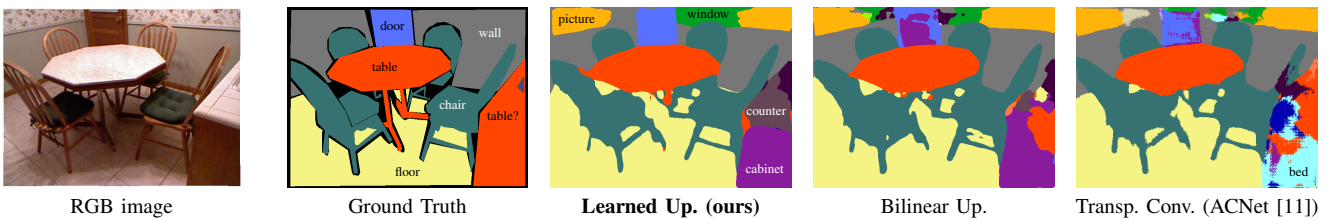

\subsection{Decoder}
As shown in Fig~\ref{fig:model}, our decoder is comprised of three decoder modules (depicted in red in Fig~\ref{fig:model}).
Our decoder module extends the one of SwiftNet~\cite{SwiftNet}, which is comprised of a $3{\times}3$ convolution with a fixed number of 128 channels and a subsequent bilinear upsampling.
However, our experiments show that for indoor RGB-D segmentation a more complex decoder is required.
Therefore, we use 512 channels in the first decoder module and decrease the number of channels in each $3{\times}3$ convolution as the resolution increases.
Moreover, we incorporate three additional Non-Bottleneck-1D-blocks to further increase segmentation performance.

Finally, we upsample the feature maps by a factor of~2.
We do not use transposed convolutions for upsampling as they are computationally expensive and often introduce undesired gridding artifacts to the final segmentation, as shown in Fig.~\ref{fig:upsampling}~(right).
Moreover, instead of using bilinear interpolation, we propose a novel light-weight learned upsampling method (see Fig.~\ref{fig:model} dark green), which achieves better segmentation results: In particular, we first use nearest neighbor upsampling to enlarge the resolution. 
Afterwards, a $3{\times}3$ depthwise convolution is applied to combine adjacent features.
We initialize the kernels such that the whole learned upsampling initially mimics bilinear interpolation.
However, our network is able to adapt the weights during training and, thus, can learn how to combine adjacent features in a more useful manner, which improves segmentation performance.

Although being upscaled, the resulting feature maps still lack fine-grained details that were lost during downsampling in the encoders.
Therefore, we design skip connections from encoder to decoder stages of the same resolution.
To be precise, we take the fused RGB-D encoder feature maps, project them with a $1{\times}1$ convolution to the same number of channels used in the decoder, and add them to the decoder feature maps.
Incorporating these skip connections results in more detailed semantic segmentations.

Similar to~\cite{SwiftNet, DeepLabv3plus}, we only process feature maps in the decoder until they are $4{\times}$ smaller than the input images and use a $3 {\times} 3$ convolution to map the features to the classes of the respective dataset.
Two final learned upsampling modules restore the resolution of the input image.

Instead of calculating the training loss only at the final output scale, we add supervision to each decoder module.
At each scale, a $1{\times}1$ convolution computes a segmentation of a smaller scale, which is supervised by the down-scaled ground truth segmentation.

\section{Experiments}
\label{sec:experiments}
We evaluate our approach on two commonly used RGB-D indoor datasets, namely SUNRGB-D~\cite{SUNRGBD} and NYUv2~\cite{NYUv2} and present an ablation study to essential parts of our network.
In order to demonstrate that our approach is suitable for other areas of application as well, we also show results on the Cityscapes~\cite{Cityscapes} dataset, the most widely used outdoor dataset for semantic segmentation.
Finally, instead of reporting benchmark results only, we present qualitative results when using our approach in a robotic indoor application.

\subsection{Implementation Details \& Datasets}
\label{sec:experiments:implementation_datasets}
We trained our networks using PyTorch~\cite{pytorch} for 500 epochs with batches of size 8.
For optimization, we used both SGD with momentum of $0.9$ and Adam \cite{kingma2015adam} with learning rates of $\{0.00125, 0.0025, 0.005, 0.01, 0.02, 0.04\}$ and $\{0.0001, 0.0004\}$, respectively, and a small weight decay of $0.0001$.
We adapted the learning rate using PyTorch's one-cycle learning rate scheduler.
To further increase the number of training samples, we augmented the images using random scaling, cropping, and flipping. 
For RGB images, we also applied slight color jittering in HSV space.

The best models were chosen based on the mean intersection over union~(mIoU).
We used bilinear upsampling to rescale the resulting class mapping to the size of the ground truth segmentation before computing the argmax for the final segmentation mask.

\textit{NYUv2 \& SUNRGB-D:} \enspace %
NYUv2 contains 1,449~indoor RGB-D images, of which~795 are used for training and 654 for testing.
We used the common 40-class label setting. 
SUNRGB-D has 37~classes and consists of 10,335~indoor RGB-D images, including all images of NYUv2.
There are 5,285~training and 5,050~testing images.
Our ablation study is based on NYUv2 as it is smaller and, thus, leads to faster trainings.
However, according to~\cite{SmallDataBigDecisions}, training on a subset is sufficient for a reliable model selection.
For both datasets, we used a network input resolution of~$640{\times}480$ and applied median frequency class balancing~\cite{MedianFrequency}.
As the input to the context module has a resolution of~$20{\times}15$ due to the downsampling of~32, we used $b=2$ branches, one with global average pooling and one with a pooling size of~$4{\times}3$.

\textit{Cityscapes:} \enspace
This dataset contains 5,000 images with fine-grained annotation for 19 classes.
The images have a high resolution of $2048{\times}1024$.
There are 2,975 images for training, 500 for validation, and 1,525 for testing.
Cityscapes also provides 20k coarsely annotated images, which \textit{we did not use for training}.
We computed corresponding depth images from the disparity images.
Since we set the network input resolution to $1024{\times}512$, the input to our context module has a resolution of $32{\times}16$, which allows $b=4$ branches in the context module, one with global average pooling and the others with pooling sizes of $16{\times}8$, $8{\times}4$, and $4{\times}2$.

For further details and other hyperparameters, we refer to our implementation available on GitHub.
\begin{figure}[t!]%
	\vspace{2mm}
    \centering
    \includegraphics[width=0.95\columnwidth]{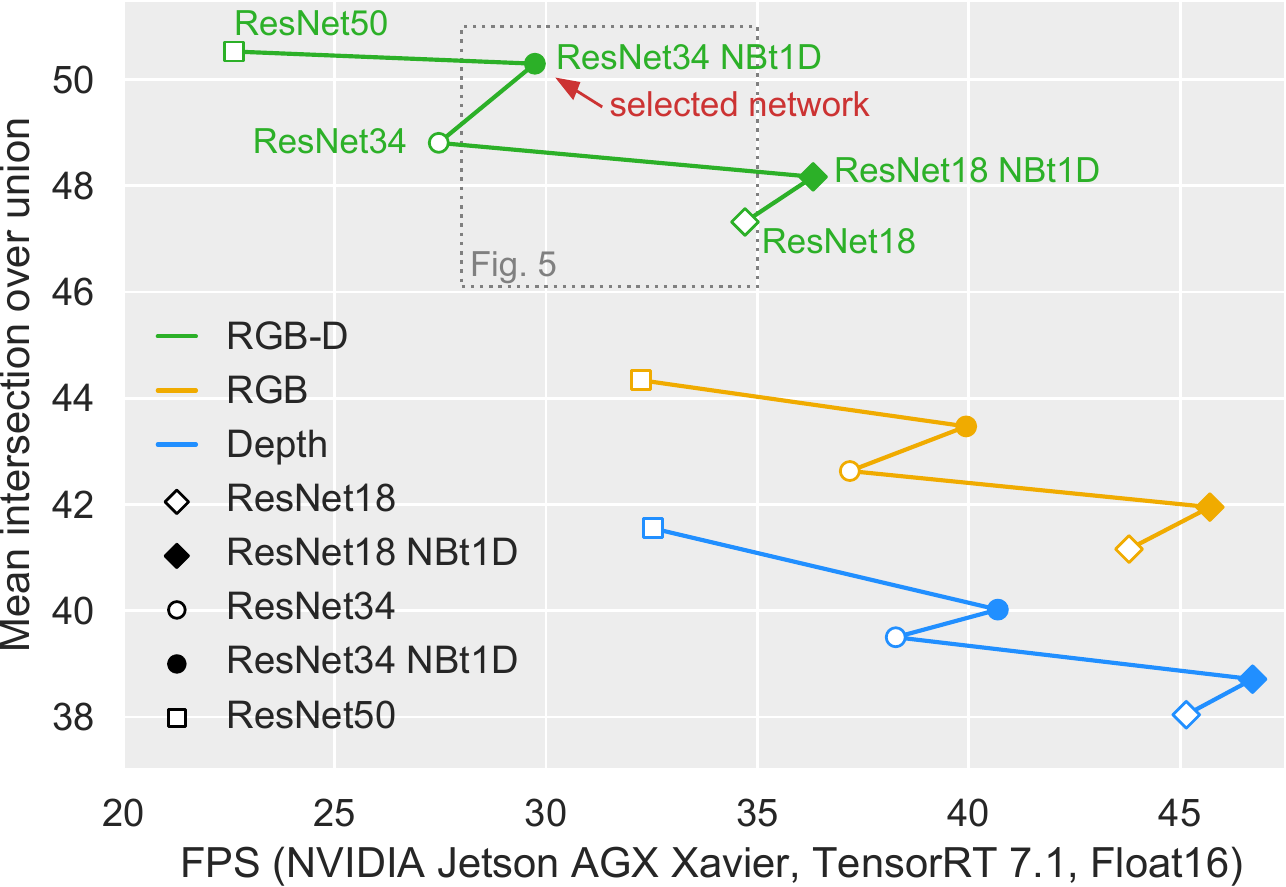}%
    
    \vspace{-2mm}
	\caption{Comparison of RGB-D to RGB and depth networks (single encoder) and different backbones on NYUv2 test set.\\[-8mm]}%
	\label{fig:exp_rgbd}
\end{figure}%

\subsection{Results on NYUv2 \& SUNRGB-D}
\label{sec:experiments:nyuv2_sunrgbd}
Fig.~\ref{fig:exp_rgbd} compares our RGB-D approach on NYUv2 to single-modality baselines for RGB and depth (single encoder) as well as evaluates different encoder backbones.
As expected, neither processing depth data nor RGB data alone reach the segmentation performance of our proposed RGB-D network.
Remarkably, the shallow ResNet18-based RGB-D network performs better than the much deeper ResNet50-based RGB network while still being faster.
Moreover, replacing ResNet's basic block with Non-Bottleneck-1D~(NBt1D) block can further improve both segmentation and inference time.
Note that ResNet50 incorporates bottleneck blocks, which cannot be replaced the same way.

Tab.~\ref{tab:sota_nyuv2_sunrgbd} lists the results of our RGB-D approach for both indoor datasets.
For the larger SUNRGB-D dataset, a similar trend can be observed.
Compared to the state of the art, our smaller ESANet achieves similar segmentation results as the often much deeper networks.
Besides focusing on segmentation performance alone, we also strive for low inference time on the embedded hardware of our robots.
Therefore, we measured the inference time for all available approaches on a NVIDIA Jetson AGX Xavier using NVIDIA TensorRT.
For our carefully designed ESANet, NVIDIA TensorRT enables up to $5{\times}$ faster inference compared to PyTorch.
As shown in Tab.~\ref{tab:sota_nyuv2_sunrgbd}~(last column), our approach enables much faster inference while performing on par or even better than other approaches.
For our application, we choose ESANet with ResNet34 backbone and Non-Bottleneck-1D~(NBt1D) block (printed in bold in Tab.~\ref{tab:sota_nyuv2_sunrgbd}) as it offers the best trade-off between inference time and performance.
The last row in Tab.~\ref{tab:sota_nyuv2_sunrgbd} further indicates that additional pretraining on synthetic data, such as SceneNet~\cite{SceneNet_RGB-D}, should be preferred to deeper backbones, especially if the target dataset is small.
\begin{figure}[t!]%
    \vspace{1.5mm}
    \centering
    \includegraphics[width=0.95\columnwidth]{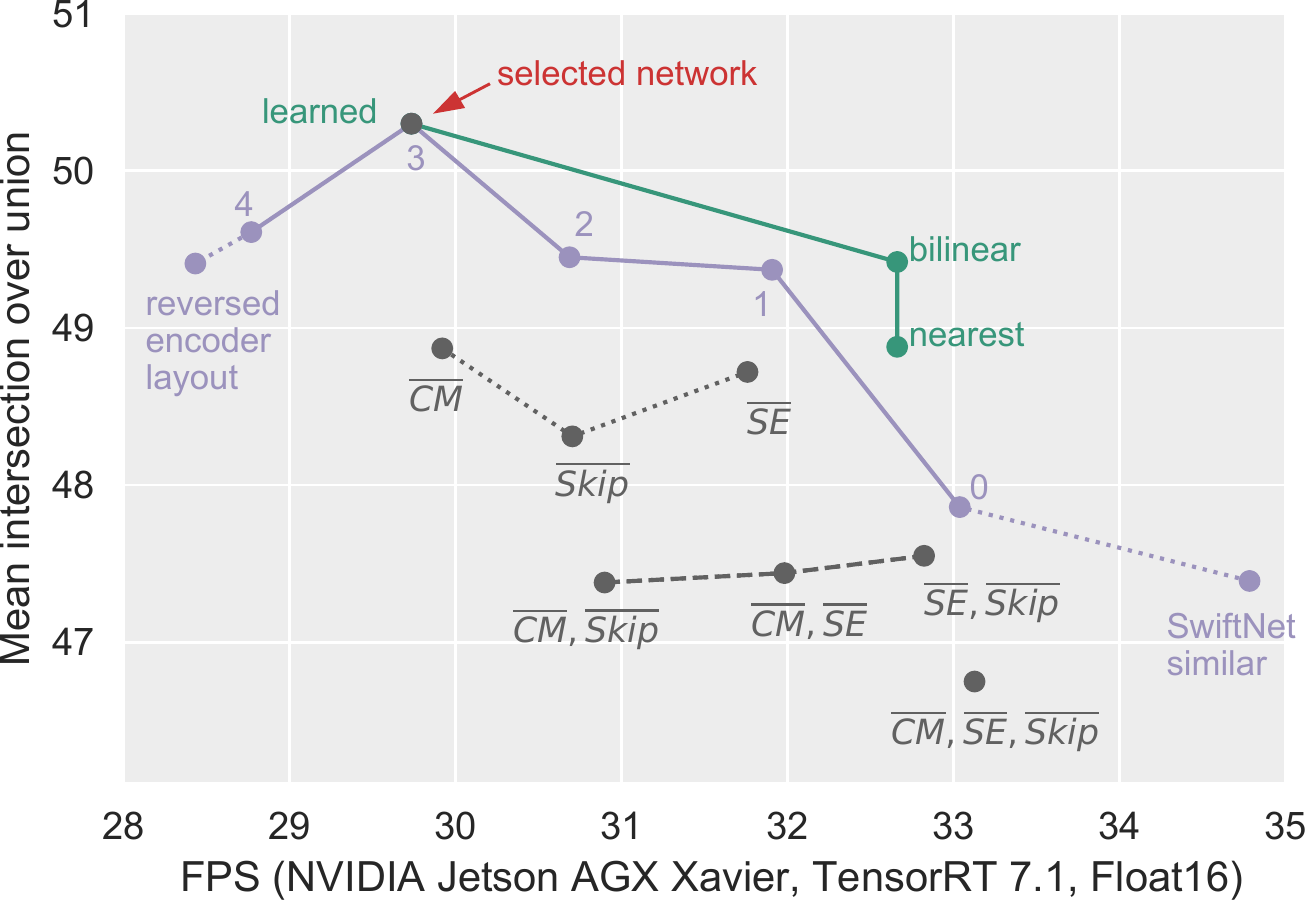}%
    
    \vspace{-1mm}
	\caption{Ablation Study on NYUv2 test. 
	Each color indicates modifying one aspect: purple: number of NBt1D blocks in decoder module, dark green: upsampling method, and gray: usage of specific network parts with {\small $\overline{CM}$}: no context module, {\small $\overline{Skip}$}: no encoder-decoder skip connections, and {\small $\overline{SE}$}: no Squeeze-and-Excitation before fusing RGB and depth.\\[-5mm]}%
	\label{fig:ablation}
\end{figure}%

\subsection{Ablation Study on NYUv2}
\label{sec:experiments:nyuv2_ablation}
Fig.~\ref{fig:ablation} shows the ablation study for fundamental parts of our network architecture and justifies our design choices.
Furthermore, it indicates the impact of each part when it is necessary to adapt our selected network to deviating real-time requirements.

As shown in purple, a shallow decoder similar to SwiftNet~\cite{SwiftNet} is not as good as more complex decoders.
Therefore, we gradually increased the number of additional NBt1D blocks in the decoder module.
Apparently, a fixed number of three blocks in each decoder module performs better than a different number or a reversed layout of the encoder's design.

In dark green, different upsampling methods in the decoder are displayed.
Although increasing inference time, the learned upsampling improves mIoU by 0.9.
Moreover, as shown in Fig.~\ref{fig:upsampling}, the obtained segmentation contains more fine-grained details compared to using bilinear interpolation.
It further prevents gridding artifacts introduced by transposed convolutions as used in ACNet~\cite{ACNet} or RedNet~\cite{RedNet}.

As shown in gray in Fig.~\ref{fig:ablation}, a context module, encoder-decoder skip connections, as well as reweighting modality-specific features with Squeeze-and-Excitation before fusion, independently improve segmentation performance.
Incorporating all three network parts leads to the best result.

\subsection{Results on Cityscapes}
\label{sec:experiments:cityscapes}
To demonstrate that our approach is applicable to other areas such as outdoor environments as well, in Tab.~\ref{tab:results_cityscapes}, we further present an evaluation on the Cityscapes dataset.

We first focus on the smaller resolution of $1024{\times}512$ as it is commonly used for efficient segmentation.
Moreover, since most approaches rely on RGB as input solely, we start by comparing a single-modality RGB version of our approach.
Efficient approaches with custom architectures such as ERFNet~\cite{ERFNet}, LEDNet~\cite{LEDNet}, and ESPNetv2~\cite{ESPNetv2} are quite fast but also perform notably worse than our ESANet.
Compared to ERFNet, LEDNet, and ESPNetv2, SwiftNet~\cite{SwiftNet} is both faster and achieves higher mIoU.
Nevertheless, with an input resolution of $1024{\times}512$, our ESANet-R34-NBt1D still exceed 30 FPS while outperforming all other efficient approaches by at least 2.2 mIoU.
Incorporating depth further increases segmentation performance.
However, the performance gain is not as high as for the indoor dataset NYUv2.
We assume that this can be deduced to the fact that the disparity images of Cityscapes are not as precise as the indoor depth images of NYUv2 and \mbox{SUNRGB-D}.
Compared to the RGB-D approach LDFNet~\cite{LDFNet} with similar inference time, we achieve notably higher mIoU.

For completeness, we also evaluated our networks on the full resolution of $2048{\times}1024$.
Compared to other methods, our ESANet lies in between mobile~(SwiftNet, BiSeNet) and non-mobile approaches for both mIoU and inference time.

However, compared to SwiftNet~(RGB, $2048{\times}1024$), our ESANet-R34-NBt1D achieves similar segmentation performance and slightly faster inference while processing RGB-D inputs with the smaller input resolution of $1024{\times}512$.

\setlength{\tabcolsep}{3pt}
\begin{table}[t]
\vspace{2mm}
\footnotesize
\begin{tabular}{@{}llllc@{}}
\toprule[1pt]
\bf Method & \bf Backbone & \bf NYUv2 & \begin{tabular}[c]{@{}c@{}} \bf SUN-\\ \bf RGB-D\end{tabular} & \multicolumn{1}{r@{}}{\bf FPS} \\ \midrule[1pt]
FuseNet \cite{FuseNet} & $2{\times\,}$VGG16 & - & 37.29 & $\dagger$ \\
\midrule%
RedNet \cite{RedNet} & $2{\times\,}$R34 & - & 46.8 & 26.0 \\ \midrule%
SSMA \cite{SSMA} & $2{\times\,}$mod. R50 & - & 44.43 & 12.4 \\
MMAF-Net \cite{MMAF-Net} & $2{\times\,}$R50 & - & 45.5 & N/A \\
RedNet \cite{RedNet} & $2{\times\,}$R50 & - & 47.8 & 22.1 \\
RDFNet \cite{RDFNet} & $2{\times\,}$R50 & 47.7* & - & 7.2 \\
ACNet \cite{ACNet} & $3{\times\,}$R50 & 48.3 & 48.1 & 16.5 \\
SA-Gate \cite{SA-Gate} & $2{\times\,}$R50 & 50.4 & 49.4* & 11.9 \\
\midrule%
SGNet \cite{SGNet} & R101 & 49.0 & 47.1 & N/A~$\triangledown$ \\
Idempotent \cite{CouplingTwoStream} & $2{\times\,}$R101 & 49.9 & 47.6 & N/A~$\triangledown$ \\
2.5D Conv \cite{2.5DConvolutionRGBD} & R101 & 48.5 & 48.2 & N/A~$\triangledown$ \\ \midrule
MMAF-Net \cite{MMAF-Net} & $2{\times\,}$R152 & 44.8 & 47.0 & N/A~$\triangledown$ \\
RDFNet \cite{RDFNet} & $2{\times\,}$R152 & 50.1* & 47.7* & 5.8 \\ \midrule[1pt]%
ESANet-R18 & $2{\times\,}$R18 & 47.32 & 46.24 & 34.7 \\
ESANet-R18-NBt1D & $2{\times\,}$R18 NBt1D & 48.17 & 46.85 & 36.3 \\
ESANet-R34 & $2{\times\,}$R34 & 48.81 & 47.08 & 27.5 \\
\textbf{ESANet-R34-NBt1D} & $2{\times\,}$R34 NBt1D & 50.30 & 48.17 & 29.7 \\
ESANet-R50 & $2{\times\,}$R50 & 50.53 & 48.31 & 22.6 \\ \midrule
ESANet (pre. SceneNet) & $2{\times\,}$R34 NBt1D & 51.58 & 48.04 & 29.7 \\
\bottomrule[1pt]
\end{tabular}

\vspace{-2mm}
\caption{Mean intersection over union of our ESANet compared to state-of-the-art methods on NYUv2 and SUNRGB-D test set ordered by SUNRGB-D performance and backbone complexity. FPS is reported for NVIDIA Jetson AGX Xavier (Jetpack 4.4, TensorRT 7.1, Float16). Legend: R:~ResNet, *:~additional test-time augmentation, i.e., flipping or multi-scale (not timed), {\small N/A}:~no implementation available, $\dagger$:~includes operations, which are not supported by TensorRT, and $\triangledown$:~expected to be slower due to complex backbone.\\[-8mm]
}
\label{tab:sota_nyuv2_sunrgbd}
\end{table}

\begin{table}[t]
\vspace{2mm}
\footnotesize
\begin{tabular}{@{}ll@{\hspace{4mm}}c@{\hspace{1.5mm}}c@{\hspace{3mm}}c@{\hspace{6mm}}c@{\hspace{1.5mm}}c@{\hspace{3mm}}c@{}}
\toprule[1pt]
 &  & \multicolumn{3}{c}{\textbf{$\mathbf{1024{\times}512}$}\mbox{\hspace{6mm}}} & \multicolumn{3}{c}{\textbf{$\mathbf{2048{\times}1024}$}} \\
 & \multicolumn{1}{l}{\textbf{Method}} & \textbf{Val} & \textbf{Test}$^{\diamond}$ & \textbf{FPS} & \textbf{Val} & \textbf{Test}$^{\diamond}$ & \textbf{FPS} \\ \midrule[1pt]
\multirow{10}{*}{\textbf{\rotatebox{90}{RGB}}} & ERFNet~\cite{ERFNet} & - & 69.7 & 49.9 & - & - & - \\
 & LEDNet~\cite{LEDNet} & - & 70.6* & 38.5 & - & - & - \\
 & ESPNetv2~\cite{ESPNetv2} & 66.4 & 66.2 & 47.4 & - & - & - \\
 & SwiftNet~\cite{SwiftNet} & 70.2 & - & 64.5 & 75.4 & 75.5 & 20.8 \\
 & BiSeNet~\cite{BiSeNet} & - & - & - & 74.8 & 74.7 & 20.0 \\
 & PSPNet~\cite{PSPNet} & - & - & - & - & 81.2* & 1.8 \\
 & DeepLabv3~\cite{DeepLabv3} & - & - & - & 79.3 & 81.3* & 0.9 \\ \cmidrule(l){2-8} 
 & ESANet-R18-NBt1D & 71.48 & - & 37.2 & 77.95 & - & 9.8 \\
 & \textbf{ESANet-R34-NBt1D} & 72.70 & 72.87 & 32.3 & 78.47 & 77.56 & 8.3 \\
 & ESANet-R50 & 73.88 & - & 24.9 & 79.23 & - & 6.5 \\ \midrule[1pt]
\multirow{6}{*}{\textbf{\rotatebox{90}{RGB-D}}} & SSMA~\cite{SSMA} & - & - & - & 82.19* & 82.31* & 2.2 \\
 & SA-Gate~\cite{SA-Gate} & - & - & - & 81.7 & 82.8 & 2.1 \\
 & LDFNet~\cite{LDFNet} & 68.48 & 71.3 & 25.3 & - & - & - \\ \cmidrule(l){2-8} 
 & ESANet-R18-NBt1D & 74.65 & - & 28.9 & 79.25 & - & 7.6 \\
 & \textbf{ESANet-R34-NBt1D} & 75.22 & 75.65 & 23.4 & 80.09 & 78.42 & 6.2 \\
 & ESANet-R50 & 75.66 & - & 16.9 & 79.97 & - & 4.0 \\ \bottomrule[1pt]
\end{tabular}
\vspace{-1mm}
\caption{Mean intersection over union of our ESANet on Cityscapes for both common input resolutions compared to state-of-the-art methods. FPS is reported for NVIDIA Jetson AGX Xavier (Jetpack 4.4, TensorRT 7.1, Float16). Legend: $\diamond$: test server result, *:~trained with additional coarse data.\\[-7mm]}
\label{tab:results_cityscapes}
\end{table}

\subsection{Application on our Robots}
\label{sec:experiments:application}
\begin{figure}[b!]%
    \vspace{-4mm}
    \definecolor{Wall}{RGB}{119,119,119}%
    \definecolor{Floor}{RGB}{244,243,131}%
    \definecolor{Ceiling}{RGB}{255,190,190}%
    \definecolor{Lamp}{RGB}{255,255,0}%
    \definecolor{Chair}{RGB}{54,114,113}%
    \definecolor{Table}{RGB}{255,69,0}%
    \definecolor{Sofa}{RGB}{0,0,176}%
    \definecolor{Picture}{RGB}{255,180,10}%
    \definecolor{Box}{RGB}{87,64,34}%
    \definecolor{TV}{RGB}{192,79,212}%
    \definecolor{Bag}{RGB}{12,83,45}%
    \definecolor{Pillow}{RGB}{24,209,255}%
    \begin{tikzpicture}[scale=0.7][every node/.style={inner sep=0,outer sep=0}]%
        \node[anchor = north east] at(0, 0.2){%
            \includegraphics[width=0.65\textwidth]{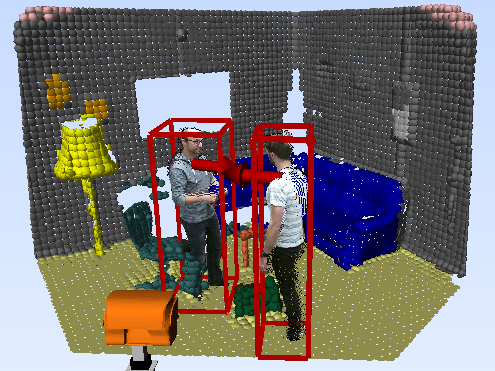}%
        };%
        \node[anchor=north east] at(-0.1, 0.1){%
            \includegraphics[width=0.21\textwidth]{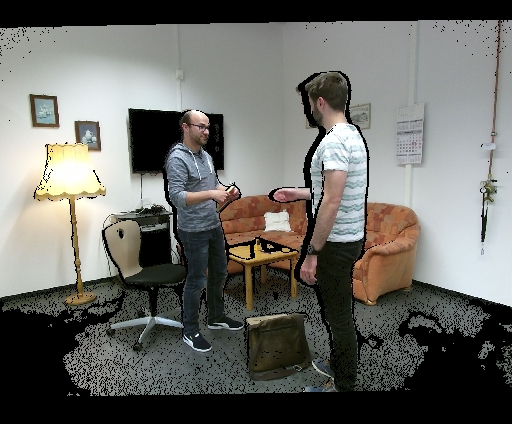}%
        };%
        \fill[Wall] (0, -0.0) rectangle (0.6, -0.4);%
        \node[anchor = west] at (0.6, -0.2){\scriptsize{Wall}};%
        \fill[Floor] (0, -0.5) rectangle (0.6, -0.9);%
        \node[anchor = west] at (0.6, -0.7){\scriptsize{Floor}};%
        \fill[Ceiling] (0, -1.0) rectangle (0.6, -1.4);%
        \node[anchor = west] at (0.6, -1.2){\scriptsize{Ceiling}};
        \fill[Lamp] (0, -1.5) rectangle (0.6, -1.9);%
        \node[anchor = west] at (0.6, -1.7){\scriptsize{Lamp}};%
        \fill[Chair] (0, -2.0) rectangle (0.6, -2.4);%
        \node[anchor = west] at (0.6, -2.2){\scriptsize{Chair}};%
        \fill[Table] (0, -2.5) rectangle (0.6, -2.9);%
        \node[anchor = west] at (0.6, -2.7){\scriptsize{Table}};%
        \fill[Sofa] (0, -3.0) rectangle (0.6, -3.4);%
        \node[anchor = west] at (0.6, -3.2){\scriptsize{Sofa}};%
        \fill[Picture] (0, -3.5) rectangle (0.6, -3.9);%
        \node[anchor = west] at (0.6, -3.7){\scriptsize{Picture}};%
        \fill[Box] (0, -4) rectangle (0.6, -4.4);%
        \node[anchor = west] at (0.6, -4.2){\scriptsize{Box}};%
        \fill[TV] (0, -4.5) rectangle (0.6, -4.9);%
        \node[anchor = west] at (0.6, -4.7){\scriptsize{TV}};%
        \fill[Bag] (0, -5) rectangle (0.6, -5.4);%
        \node[anchor = west] at (0.6, -5.2){\scriptsize{Bag}};%
        \fill[Pillow] (0, -5.5) rectangle (0.6, -5.9);%
        \node[anchor = west] at (0.6, -5.7){\scriptsize{Pillow}};%
    \end{tikzpicture}
    
    \vspace{-2mm}
    \caption{Application in our robotic scene analysis system.}
    \label{fig:application}
\end{figure}
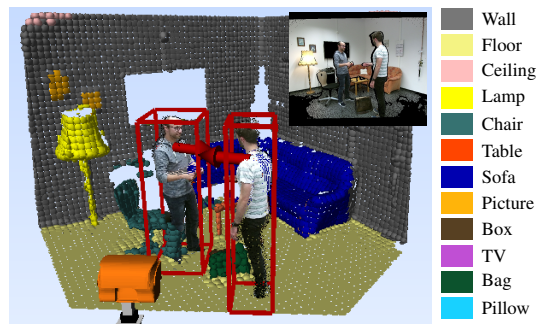

Instead of evaluating on benchmark datasets only, we further present qualitative results with a~Kinect2 sensor~\cite{freenect2, freenect2_kde} in one of our indoor applications.
We deployed our proposed ESANet-R34-NBt1D to our robot in order to accomplish the complex system for semantic scene analysis shown in Fig.~\ref{fig:eyecatcher}.
The obtained segmentation masks enrich the robot's visual perception enabling stronger person perception and robust semantic mapping including a refined floor representation which indicates free space. 
Fig.~\ref{fig:application} provides an insight into the entire system. 
For further qualitative results and a comparison to non-semantic scene perception, we refer to the attached video or our repository on GitHub.

\section{Conclusion}
\label{sec:conclusion}
In this paper, we have presented an efficient \mbox{RGB-D} segmentation approach, called ESANet, which is characterized by two enhanced ResNet-based encoders utilizing the \mbox{Non-Bottleneck-1D} block, an attention-based fusion for incorporating depth information, and a decoder utilizing a novel learned upsampling. 
On the indoor datasets NYUv2 and SUNRGB-D, our ESANet performs on par or even better while enabling much faster inference compared to other state-of-the-art methods.
Thus, it is well suited for embedding in a complex system for scene analysis on mobile robots given limited hardware.

\bibliographystyle{bib/IEEEtran}
\bibliography{bib/literature}

\end{document}